\documentclass{article}
\usepackage[nonatbib, final]{nips_style}
\usepackage[utf8]{inputenc} 
\usepackage[T1]{fontenc}    
\usepackage{hyperref}       
\usepackage{url}            
\usepackage{booktabs}       
\usepackage{amsfonts}       
\usepackage{nicefrac}       
\usepackage{microtype}      
\usepackage{float}
\usepackage{subcaption}
\usepackage{graphicx}
\usepackage{enumitem}
\usepackage[table,xcdraw]{xcolor}
\usepackage{adjustbox}
\usepackage{hyperref}
\usepackage{svg}
\usepackage{sectsty}
\usepackage{titlesec}
\usepackage{url}
\usepackage{float}
\usepackage{microtype}
\usepackage{graphicx} 
\usepackage{amsmath}
\usepackage[utf8]{inputenc} 
\usepackage[T1]{fontenc}    
\usepackage{hyperref}       
\usepackage{url}            
\usepackage{booktabs}       
\usepackage{amsfonts}       
\usepackage{nicefrac}       
\usepackage{microtype}      
\usepackage{xcolor}         
\usepackage{xspace}
\usepackage{amsmath}
\usepackage{amssymb}
\usepackage{booktabs}
\usepackage{makecell}
\usepackage{colortbl}
\usepackage{graphicx}   
\usepackage{CJKutf8}
\usepackage{multirow}
\usepackage{interval}
\intervalconfig{soft open fences}
\usepackage{algorithm, algpseudocode}

\usepackage{float}
\usepackage{enumitem}
\usepackage{tabularx}
\usepackage{placeins}
\usepackage{color}
\usepackage{tcolorbox}
\usepackage{verbatim}
\usepackage{colortbl} 
\usepackage{listings}
\usepackage{makecell}
\usepackage{siunitx}
\usepackage{xcolor}

\definecolor{bg}{RGB}{176,226,255}
\definecolor{bonus_green}{RGB}{0,100,0}

\title{What Makes Diffusion Language Models \\ Super Data Learners?}

\author{
\hspace{1mm}\textbf{Zitian Gao} \hspace{2mm}
\textbf{Haoming Luo} \hspace{2mm}
\textbf{Lynx Chen} \hspace{2mm}
\textbf{Jason Klein Liu} \quad \\ \\
\hspace{6mm}\textbf{Ran Tao} \hspace{3mm}
\textbf{Joey Zhou} \hspace{2mm}
\textbf{Bryan Dai}\thanks{~Corresponding author.} \vspace{3mm} \\
\hspace{6mm} Ubiquant \quad \vspace{2mm} \\
\hspace{1mm}\texttt{\{ztgao02,hmluo,ylchen,rbliu,rtao02,jzhou,cbdai\}@ubiquant.com} \\
}

\begin{document}
\maketitle

\begin{abstract}
Recent studies have shown that diffusion language models achieve remarkable data efficiency under limited-data constraints, yet the underlying mechanisms remain unclear. In this work, we perform extensive ablation experiments to disentangle the sources of this efficiency. Our results show that random masking of input tokens plays the dominant role. We further show that similar gains can be obtained through dropout in attention and MLP layers, as well as weight decay, indicating that stochastic regularization broadly enhances data efficiency in multi-epoch training. Our code is available at \url{https://github.com/zitian-gao/data-efficiency}.


\end{abstract}

\section{Introduction}

The token crisis in large language model (LLM) pretraining has become a widely acknowledged challenge \cite{runout,torepeat,muennighoff2025scalingdataconstrainedlanguagemodels,jinjie,mihir}. Although the internet contains the collective knowledge of humanity, the supply of high-quality pretraining tokens is reaching a plateau \cite{runout}, imposing a bottleneck on the scaling of modern LLMs. In addition to developing synthetic data generation methods to artificially expand high-quality corpora, researchers have increasingly turned their attention to improving the data efficiency of pretraining—namely, extracting more capability from fewer tokens.

Muennighoff et al.\cite{muennighoff2025scalingdataconstrainedlanguagemodels} show that looping autoregressive Transformers over the same corpus is a poor substitute for fresh data. At fixed compute, single-epoch training yields the best validation loss per FLOP, with only minor gains up to 4 epochs; beyond that, repeated tokens add little value, collapsing to near zero after dozens of epochs. Vanilla autoregressive multi-epoch training shows classic overfitting—training loss falls while validation loss rises as unique tokens dwindle—indicating memorization rather than generalization. Their data-constrained scaling laws capture this decay, suggesting that when repetition is unavoidable, increasing epochs may justify fewer parameters, but returns diminish sharply; expending compute on repeated data is largely wasteful \cite{muennighoff2025scalingdataconstrainedlanguagemodels}.

Diffusion language models (DLMs) have recently emerged as a promising direction in this context. Empirical studies suggest that DLMs can achieve competitive or even superior downstream performance under severely limited data budgets \cite{jinjie,mihir}, yet the mechanisms underlying this surprising efficiency remain poorly understood. Without a clear understanding of why DLMs are such “super data learners,” it is difficult to evaluate their true potential or design principled improvements.

In this work, we take a systematic approach to dissecting the training dynamics of DLMs. Through a series of controlled ablation studies, we identify random input masking as the key factor driving their superior data efficiency, while also highlighting the roles of dropout, weight decay, and multi-epoch training strategies. We further benchmark our models against strong baselines such as Qwen3-0.6B, revealing that a 0.6B-parameter DLM trained on only 3B unique tokens can outperform a counterpart trained on 36T tokens. These findings not only shed light on the mechanisms behind DLMs but also provide concrete strategies for building more data-efficient LLMs in the face of the token crisis.

Our main contributions are as follows:
\begin{itemize}[leftmargin=*]
    \item Through extensive ablations, we demonstrate that random masking of input tokens is the key factor that makes diffusion language models super data learners, which we term as \textbf{``token dropout.''}
    
    \item We show that vanilla autoregressive Transformers, when augmented with carefully designed stochastic regularization such as MLP dropout and weight decay, can also exhibit the ``super data learner'' behavior, and in some cases even outperform diffusion language models.
    
    
\end{itemize}

\section{Preliminaries}

\subsection{Auto-regressive Model}
Auto-regressive~(AR) language models directly model the joint probability of a sequence $\mathbf{x} = (x_1, x_2, \dots, x_n)$ via the chain rule:
\begin{equation*}
    p_\theta(\mathbf{x}) = \prod_{t=1}^n p_\theta(x_t \mid x_{<t}),
\end{equation*}
where $x_{<t} = (x_1, \dots, x_{t-1})$ represents the prefix tokens.  
The training objective is maximum likelihood estimation (MLE), equivalent to minimizing the negative log-likelihood (NLL):
\begin{equation*}
    \mathcal{L}_{\text{AR}}(\theta) = - \sum_{t=1}^n \log p_\theta(x_t \mid x_{<t}).
\end{equation*}
In practice, $p_\theta(x_t \mid x_{<t})$ is parameterized by a Transformer, where each token is embedded, passed through multi-head self-attention, and projected to the vocabulary space.

\subsection{Diffusion-style Models for Language}
Diffusion-style language models define a generative distribution over discrete token sequences via a forward masking process and a learned reverse denoising process. Let $x_0=(x_0^1,\dots,x_0^L)$ be a clean sequence from the data distribution. The forward process draws a masking ratio $t\in[0,1]$, then constructs a partially observed sequence $x_t$ by independently replacing each token with a special mask symbol $\mathrm{M}$ with probability $t$ (and keeping it with probability $1-t$). As $t$ approaches $1$, $x_t$ becomes fully masked; as $t$ approaches $0$, $x_t$ converges to $x_0$.

A non-causal Transformer parameterizes a mask predictor $p_\theta(\cdot\mid x_t)$ that observes the entire masked sequence and outputs per-position distributions over the vocabulary. Training is performed by denoising masked tokens at random corruption levels $t\sim \mathrm{Uniform}(0,1)$, using a cross-entropy loss computed only on masked positions:
\begin{equation*}
\mathcal{L}(\theta)= -\,\mathbb{E}_{x_0\sim p_{\text{data}},\, t\sim U[0,1],\, x_t}\left[\frac{1}{t}\sum_{i=1}^L \mathbf{1}[x_t^i=\mathrm{M}]\,\log p_\theta(x_0^i\mid x_t)\right].
\end{equation*}
Here, $x_t$ is obtained by masking $x_0$ at rate $t$. The factor $1/t$ normalizes the contribution of each training example by the expected number of masked tokens.

\paragraph{Training mechanics.} Each minibatch repeats the following steps: (i) sample $t\sim U[0,1]$ for every sequence; (ii) independently mask each token of $x_0$ with probability $t$ to form $x_t$; (iii) run the non-causal Transformer on $x_t$; (iv) compute cross entropy only at masked positions and backpropagate. Unlike masked language modeling with a fixed masking rate, sampling $t$ across $[0,1]$ exposes the model to a full spectrum of corruption levels, aligning the training distribution with the discretized reverse process used at inference. Standard LLM components (optimizer, scheduling, dropout, weight decay) apply unchanged; KV caching is not used because the model is non-causal and conditions on the full masked input.

\paragraph{Inference (sampling).} Generation is implemented by discretizing the reverse process with $K$ steps. Given a prompt $p_0$ and a target length $L$, choose a decreasing schedule $1=t_K>\cdots>t_1>t_0=0$.
\begin{enumerate}
\item Initialize $r_{t_K}$ as a fully masked sequence of length $L$.
\item For $k=K,\dots,1$:
  \begin{enumerate}
  \item Predict token distributions at masked positions using $p_\theta(\cdot\mid p_0, r_{t_k})$ and fill them by sampling (or argmax).
  \item Remask a fraction $\rho_k \approx t_{k-1}/t_k$ of positions to obtain $r_{t_{k-1}}$, so that the expected masking level matches the forward process at $t_{k-1}$.
  \end{enumerate}
\item Output $r_{t_0}$ as the final sample.
\end{enumerate}
Remasking can be purely random (theoretically faithful) or use practical heuristics: (i) low-confidence remasking (drop the least confident predictions to focus denoising where uncertainty is high), and (ii) semi-autoregressive block updates (partition the sequence into blocks, proceed left-to-right across blocks while fully denoising within each block). More steps generally improve quality at higher compute cost; uniform or cosine schedules are commonly used and easy to implement.

\paragraph{Summary.} Diffusion-style language models train a non-causal denoiser with random masking levels $t\sim U[0,1]$ and perform inference via a discretized reverse process with prediction and remasking. This yields a principled generative model over sequences, compatible with standard LLM training pipelines while replacing autoregressive decoding with iterative denoising.

\subsection{Dropout}
Dropout is a stochastic regularization technique that prevents overfitting by randomly masking neurons during training \cite{dropout}. Given a hidden vector $\mathbf{h} \in \mathbb{R}^d$, dropout applies a random binary mask $\mathbf{m} \sim \text{Bernoulli}(p)^d$:
\begin{equation*}
    \tilde{\mathbf{h}} = \frac{1}{1-p}\, (\mathbf{m} \odot \mathbf{h}),
\end{equation*}
where $p$ is the dropout rate and $\odot$ denotes element-wise multiplication. The scaling factor $1/(1-p)$ ensures expectation preservation:
\begin{equation*}
    \mathbb{E}[\tilde{\mathbf{h}}] = \mathbf{h}.
\end{equation*}

\paragraph{Attention layers.} In Transformers, dropout is often applied to the attention weights:
\begin{equation*}
    \text{Attention}(\mathbf{Q},\mathbf{K},\mathbf{V}) = \text{Softmax}\!\left(\frac{\mathbf{QK}^\top}{\sqrt{d_k}}\right)\mathbf{V},
\end{equation*}
where dropout is applied to the softmax output before multiplying with $\mathbf{V}$.

\paragraph{MLP layers.} Similarly, for feed-forward networks (FFNs) with hidden activation $\mathbf{h}$:
\begin{equation*}
    \mathbf{h}' = \sigma(\mathbf{W}_1 \mathbf{h} + \mathbf{b}_1), \quad
    \mathbf{y} = \mathbf{W}_2 (\text{Dropout}(\mathbf{h}')) + \mathbf{b}_2,
\end{equation*}
where $\sigma(\cdot)$ is a nonlinearity.

\subsection{Token Dropout}
Token Dropout is a sequence-level regularization technique that randomly suppresses a subset of time steps in the input hidden states during training. Unlike feature-wise dropout that acts independently across the hidden dimension, Token Dropout operates along the temporal axis, encouraging robustness to missing or sparse context and reducing over-reliance on immediately adjacent tokens.

Let $\mathbf{H}\in\mathbb{R}^{S\times B\times d}$ denote the decoder input hidden states for sequence length $S$, batch size $B$, and hidden width $d$. In the standard formulation of token dropout, a fixed drop probability $p\in[0,1)$ is chosen and a temporal mask is sampled i.i.d. across time for each sample:
\begin{equation*}
m_{s,b} \sim \mathrm{Bernoulli}(1-p), \quad s=1,\dots,S,\ \ b=1,\dots,B,
\end{equation*}
and the masked states are $\tilde{\mathbf{H}}{s,b,:}=m{s,b},\mathbf{H}{s,b,:}$. When aligning with diffusion-style training for LLMs, we instead adopt a per-sample random drop ratio to expose the model to a spectrum of visible-context densities. Concretely, for each sample $b$ we draw
\begin{equation*}
r_b \sim \mathrm{Uniform}(0,1),
\end{equation*}
and then sample the temporal mask with a keep probability $1-r_b$:
\begin{equation*}
m{s,b} \sim \mathrm{Bernoulli}(1-r_b),\quad s=1,\dots,S.
\end{equation*}

Token dropout is applied only during training and is disabled at inference, analogously to the standard dropout. 

By default, we do not apply expectation-preserving scaling, interpreting the operation as simulating missing context rather than feature-wise dropout.

\subsection{Weight Decay}
Weight decay is a regularization method equivalent to $\ell_2$ penalty on the model parameters \cite{loshchilov2017decoupled}. For a parameter vector $\theta$, the modified loss is:
\begin{equation*}
    \mathcal{L}_{\text{WD}}(\theta) = \mathcal{L}(\theta) + \frac{\lambda}{2} \|\theta\|_2^2,
\end{equation*}
where $\lambda > 0$ is the regularization coefficient.  
In gradient descent, this yields the update rule:
\begin{equation*}
    \theta \leftarrow \theta - \eta \left( \nabla_\theta \mathcal{L}(\theta) + \lambda \theta \right),
\end{equation*}
where $\eta$ is the learning rate.

\subsection{A Unified Perspective on AR, Diffusion-Style LLMs, and Regularization}

We split the connections into: (i) modeling/training paradigms (objectives and conditioning), and (ii) regularization (capacity control). 

\subsubsection{Part I: Connecting AR, diffusion LLMs, and token dropout.}
AR decoding models $p_\theta(x_t\!\mid\!x_{<t})$ and learns to predict the next token given a strictly left context; under teacher forcing, an input prefix like ``A B C'' is trained to target the shifted sequence ``B C D'': 
\[
\textbf{AR:}\quad \underbrace{\text{A B C}}_{\text{input}} \Rightarrow \underbrace{\text{B C D}}_{\text{targets}}.
\]
Introducing \emph{token dropout} in an AR setting removes some prefix tokens but leaves the objective unchanged—predict the same next tokens—thereby augmenting the training samples.
\[
\textbf{Token drop (AR-style):}\quad \text{A [MASK] C} \Rightarrow \text{B C D}.
\]
By contrast, diffusion-style LLMs use a \emph{non-causal} denoiser that predicts masked tokens anywhere in the sequence and iteratively refines a partially observed sample. With random masking levels (akin to the corruption rate $t$), the model reconstructs the clean sequence by alternating predict–remask steps:
\[
\textbf{Diffusion LLM (non-causal + token drop):}\quad 
\text{A [MASK] C} \to \text{A B C},\quad
\text{A B [MASK]} \to \text{A B C}.
\]
Conceptually: AR predicts \emph{future from past}; AR+token-drop predicts the same future with \emph{partial past}; diffusion+token-drop reconstructs the \emph{present everywhere} from partially observed context, aligning training (random $t$) with iterative denoising at inference.

\subsubsection{Part II: Regularization (Capacity Control Orthogonal to Objectives)}
Regularization mechanisms operate independently of the training objective or conditioning scheme. They leave targets unchanged and primarily affect capacity, stability, and generalization:

\begin{description}
\item[Feature-wise Dropout.]
Randomly zeros components of hidden activations with expectation-preserving scaling, reducing co-adaptation and overfitting. Common sites include attention probabilities and MLP activations, where injected noise acts as feature-level smoothing.

\item[Token Dropout (Sequence-level).]
Suppresses entire tokens along the temporal axis. In AR models, it simulates missing left context; in diffusion-style training, it directly mirrors the corruption process. Compared to feature-wise dropout, its effect dimension shifts from \emph{features} to \emph{time steps}, better aligning with the generative process.

\item[Weight Decay.]
Applies an $\ell_2$ penalty on parameters, discouraging overly large weights and improving generalization. Decoupled formulations such as AdamW are widely used. Weight decay complements dropout: while dropout injects stochasticity into representations, weight decay constrains the magnitudes of parameters.
\end{description}

\paragraph{Practical insights.}
\begin{itemize}
\item \textbf{Unifying objectives and conditioning:} AR and diffusion-style LLMs both maximize likelihood but differ in factorization—time-based versus corruption-based. Iterative refinement serves as a computational bridge.
\item \textbf{Consistency of corruption:} Token dropout provides a common corruption lens. For AR, it ensures robustness to partial context; for diffusion-style training, it constitutes the learning signal. Training schedules over $t$ and inference budgets $K$ should be tuned jointly.
\item \textbf{Orthogonality of capacity control:} Feature-wise dropout, token dropout, and weight decay are independent of the learning objective. They should be co-tuned alongside corruption schedules and decoding strategies to achieve optimal quality–latency–robustness trade-offs under fixed compute budgets.
\end{itemize}
\section{Method}

\subsection{Model Architecture}
Our model architecture follows the design of Qwen3-0.6B \cite{yang2025qwen3technicalreport}, with modifications to support our experimental setup. 
Specifically, the model is composed of 28 transformer layers, each with a hidden size of 1024 and a feed-forward network (FFN) dimension of 3072. 
The attention mechanism adopts group query attention (GQA), with 16 total attention heads and 8 query groups. 
Each head has 128 key-value channels, and rotary position embeddings with a rotary base of one million are employed. 
To enhance stability, we use RMSNorm with $\epsilon=10^{-6}$ and apply QK-layer normalization. 
Other architectural settings include SwiGLU activations, bias-free linear projections, and the Transformer Engine backend with fused attention kernels. 
The maximum sequence length is set to 4096, while positional embeddings are extended to 40,960. 
All parameters are initialized with a standard deviation of 0.02. Bf16 precision is used during training.

\subsection{Training Settings}
We train the model using Megatron \cite{shoeybi2020megatronlmtrainingmultibillionparameter} on randomly sampled 3 billion token from the \texttt{olmo-mix-1124} corpus \cite{olmo20252olmo2furious}, comprising 3.9 trillion unique tokens. The dataset includes a mixture of multilingual, code, math, and general-domain texts. We reshuffle the dataset between epochs to ensure training diversity, and adopt a train-validation split of 99:1. The optimizer is AdamW with $\beta_1=0.9$, $\beta_2=0.95$, $\epsilon=10^{-8}$, and default weight decay of 0.1. The learning rate is set to $3\times 10^{-4}$ with 2000 warmup steps, followed by a constant learning rate schedule. We apply gradient clipping at 1.0, and use a global batch size of 2048.

One epoch consists of about 357 steps, and we train for 120 epochs to effectively compare different methods.

\subsection{Evaluation}
We primarily evaluate six downstream task capabilities of the models using two widely adopted benchmarks: HellaSwag \cite{zellers2019hellaswag}, PIQA \cite{bisk2019piqareasoningphysicalcommonsense}, SIQA \cite{siqa}, Winogrande \cite{sakaguchi2019winograndeadversarialwinogradschema}, Lambada \cite{paperno2016lambadadatasetwordprediction}, and ARC-e \cite{arc}, with evaluations conducted via OpenCompass \cite{2023opencompass}.
\section{Experiments}
We design a series of experiments to systematically analyze the data efficiency of diffusion language models (DLMs). Our focus is to disentangle the effects of input masking, stochastic regularization, and training dynamics. Unless otherwise specified, all models are trained under identical settings with comparable parameter counts.  

\subsection{Main Results}

\begin{table*}[ht]
\centering
\small
\begin{tabularx}{\textwidth}{lcccccccc}
\toprule
\textbf{Model} & \textbf{ARC-e} & \textbf{HellaSwag} & \textbf{Lambada} & \textbf{PIQA} & \textbf{SIQA} & \textbf{Winogrande} & \textbf{Average} \\
\midrule
AR & 40.04 & 39.8 & 44.05 & 66.05 & 39.25 & 51.78 & 46.83 \\
Diffusion-style Input & 41.62 & 41.16 & 47.10 & 67.08 & 38.84 & 51.78 & 47.98 \\
DLM & 42.15 & 41.19 & 47.74 & 68.06 & 38.84 & 50.51 & 48.08 \\
Weight Decay 0.5 & \textbf{44.27} & \textbf{43.59} & 50.82 & 68.28 & \textbf{40.63} & 52.09 & 49.95 \\
MLP Dropout 0.1 & 44.09 & \textbf{43.59} & \textbf{56.63} & \textbf{69.26} & 40.48 & \textbf{53.75} & \textbf{51.30} \\
\bottomrule
\end{tabularx}

\caption{Overall downstream evaluation results on six benchmarks (ARC-e, HellaSwag, Lambada, PIQA, SIQA, Winogrande). 
We compare standard Autoregressive (AR) training, Diffusion Language Model (DLM), and Diffusion-style Input training AR model, as well as ablations with regularization parameters including Weight Decay (0.5) and MLP Dropout (0.1). 
The table reports accuracy (\%) across all tasks and their average. Best results per column are highlighted in \textbf{bold}.}
\label{tab:overall-results}
\end{table*}

\subsection{Ablation of Diffusion Language Model}
We compare three setups: (1) diffusion input with diffusion loss (referred to as Full DLM in the figure), (2) diffusion input with an autoregressive (AR) loss (shown as Diffusion-style Input), and (3) standard autoregressive (AR) training (labeled AR). This ablation disentangles the effects of the diffusion loss from those of the diffusion-style input corruption.

\begin{figure}[H]
    \centering
    \includegraphics[width=1.0\linewidth]{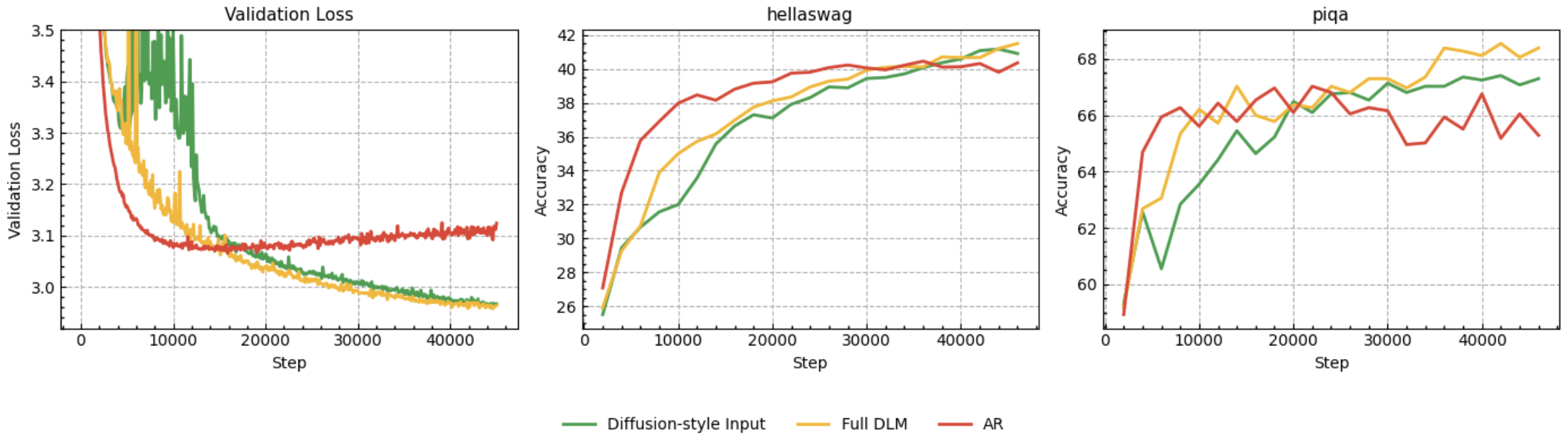}
    \caption{The validation loss of the 3 methods in the first figure on the left, and the accuracies of the 3 methods on the downstream metrics Hellaswag and PIQA in the two figures on the right.}
    \label{fig:fig1}
\end{figure}

As shown in Figure 1, the vanilla AR exhibits a rise in validation loss at around 10,000 steps, indicating overfitting, and the downstream metrics begin to decline at around 30,000 steps. In contrast, both Full DLM and Diffusion-style input maintain a steady downward trend. Moreover, the validation losses and downstream metrics of DLM and Diffusion-style Input are nearly identical, suggesting that the “super data learner” property of the Diffusion Language Model (DLM) primarily stems from the Diffusion-style Input (which we refer to as token dropout).

\subsection{Token Dropout Ratio}
We investigate the effect of the token drop ratio by evaluating models trained with ratios of 1.0 (Diffusion-style input), 0.5, 0.3, 0.1, and 0 (Vanilla AR). This experiment identifies whether extreme masking is beneficial or whether moderate masking yields superior performance.  

\begin{figure}[H]
    \centering
    \includegraphics[width=1\linewidth]{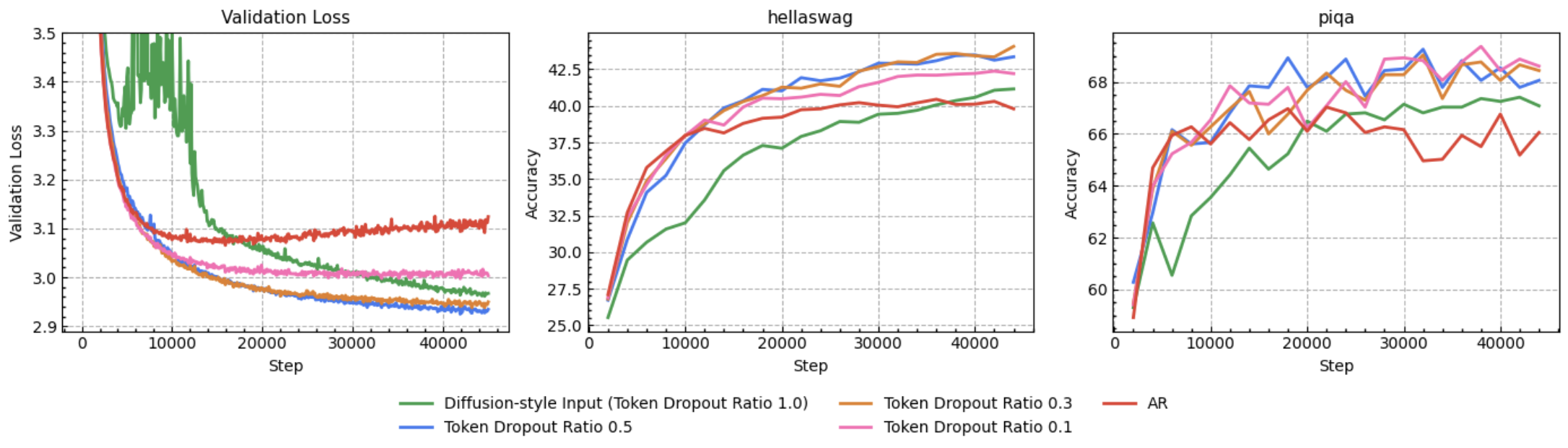}
    \caption{The validation loss of the 5 methods in the first figure on the left, and the accuracies of the 5 methods on the downstream metrics Hellaswag and PIQA in the two figures on the right.}
    \label{fig:fig2}
\end{figure}

As shown in Figure 2, a token dropout ratio of 0.1 fails to completely prevent the AR model from overfitting, whereas ratios of 0.3 and 0.5 succeed. This indicates that the threshold ratio for token dropout to prevent AR overfitting lies between 0.1 and 0.3. Moreover, although a token dropout ratio of 1.0 results in a higher absolute validation loss in the early stage, its later loss reduction rate and downstream metric improvement rate surpass those of the other ratios. Therefore, we conclude that larger token dropout ratios are more beneficial for improving data efficiency.

\subsection{Dropout in Attention}
To test whether stochastic regularization beyond input masking contributes to data efficiency, 
we compare AR training against models with dropout applied to attention layers at rates of 0.1, 0.3, and 0.5.  

\begin{figure}[H]
    \centering
    \includegraphics[width=1\linewidth]{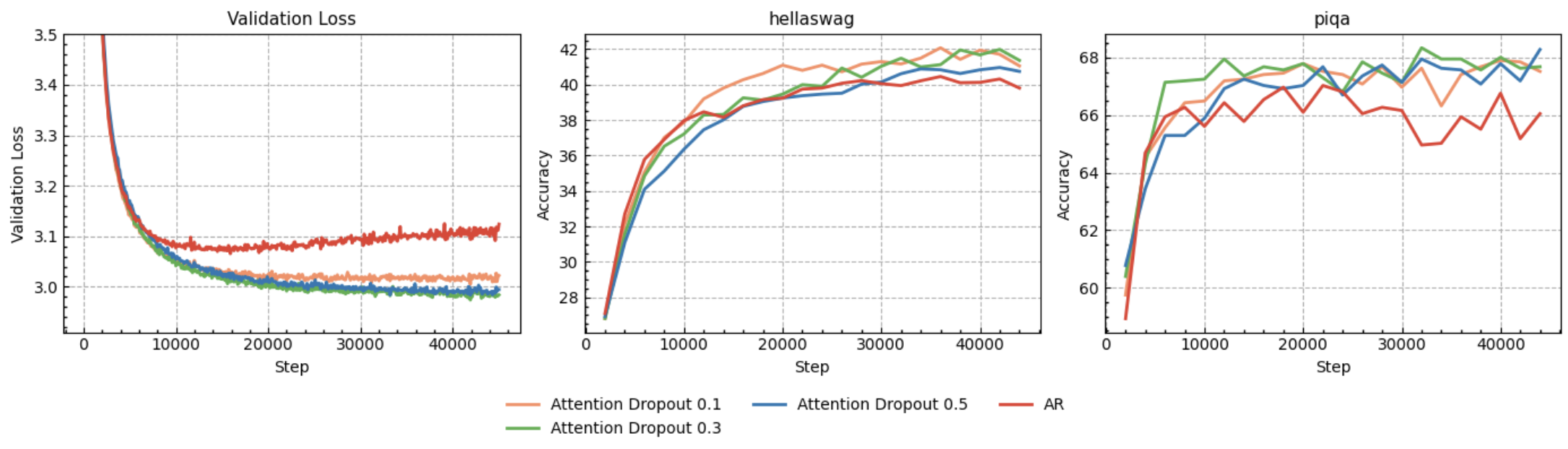}
    \caption{The validation loss of the 4 methods in the first figure on the left, and the accuracies of the 4 methods on the downstream metrics Hellaswag and PIQA in the two figures on the right.}
    \label{fig:fig3}
\end{figure}

As shown in Figure 3, even an attention dropout of 0.1 can prevent the validation loss from rising, but it fails to produce significant improvements on downstream metrics. This suggests that attention dropout may not contribute to improving data efficiency in multi-epoch training.

\subsection{Dropout in MLP}
Similarly, we compare AR training against models with dropout applied to MLP layers at rates of 0.1, 0.3, and 0.5.  

\begin{figure}[H]
    \centering
    \includegraphics[width=1\linewidth]{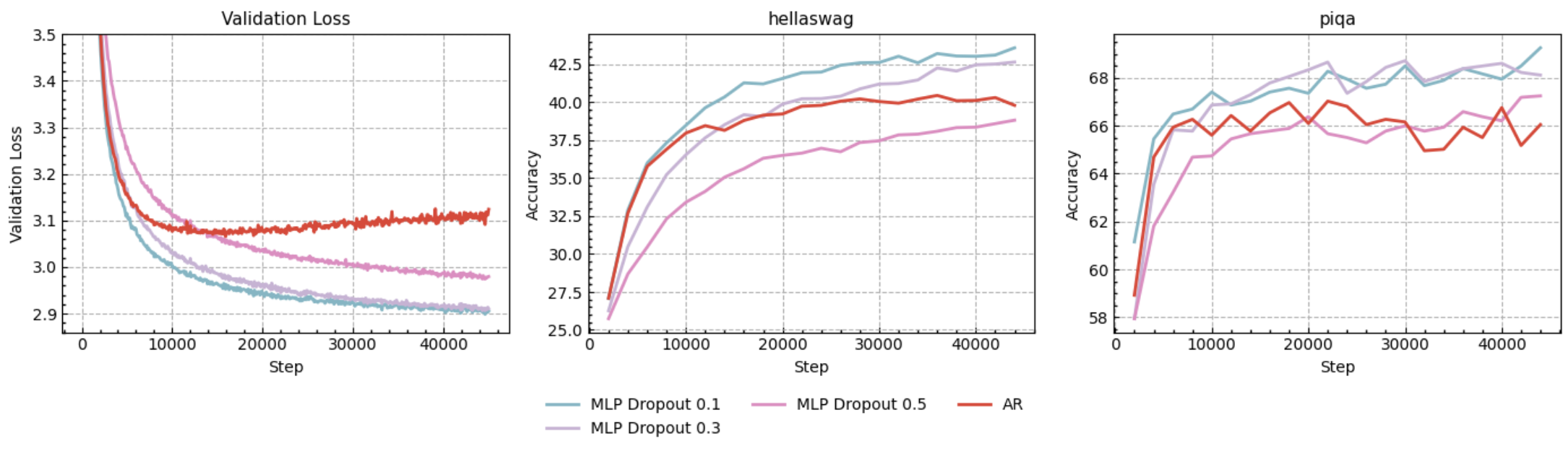}
    \caption{The validation loss of the 4 methods in the first figure on the left, and the accuracies of the 4 methods on the downstream metrics Hellaswag and PIQA in the two figures on the right.}
    \label{fig:fig4}
\end{figure}

As shown in Figure 4, the dropout effects of MLP 0.1 and 0.3 are very significant. Both the validation loss and downstream metrics are substantially better than the autoregressive baseline without dropout, whereas a 0.5 dropout rate is too high.

\subsection{Weight Decay}
We further study the role of weight decay as another form of regularization. Specifically, we compare AR models trained with weight decay values of 0.1, 0.3, and 0.5.  

\begin{figure}[H]
    \centering
    \includegraphics[width=1\linewidth]{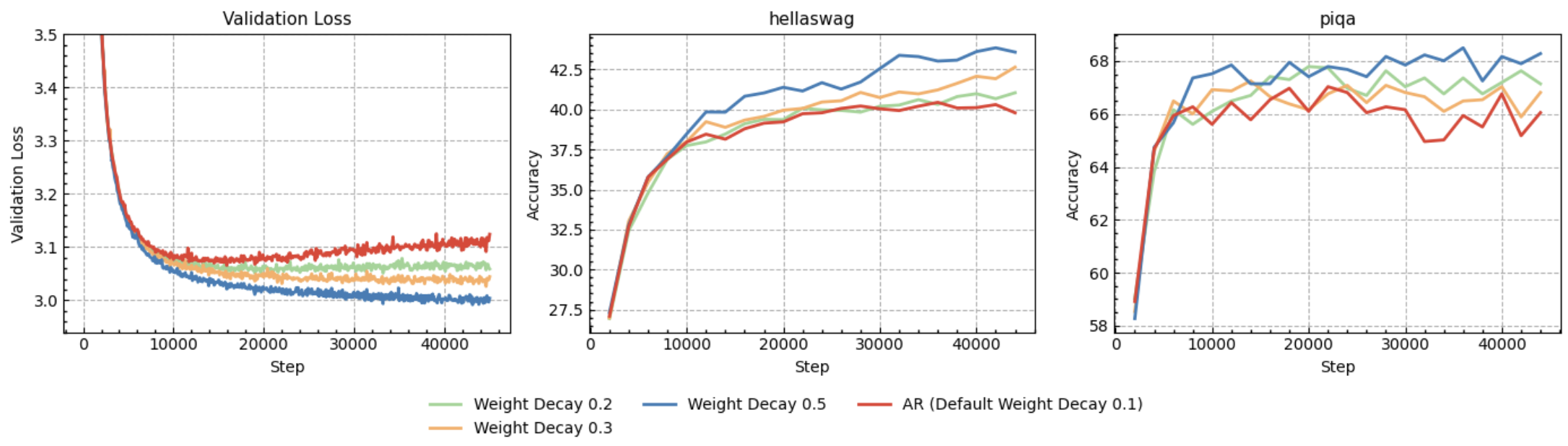}
    \caption{The validation loss of the 4 methods in the first figure on the left, and the accuracies of the 4 methods on the downstream metrics Hellaswag and PIQA in the two figures on the right.}
    \label{fig:fig5}
\end{figure}

From Figure 5, we observe that within the range of 0.1–0.5, weight decay and multi-epoch training exhibit a monotonic improvement in data efficiency. However, a weight decay of 0.1, which is commonly used as a default setting, fails to prevent overfitting in autoregressive models.





\section{Related Work}

\subsection{Data Efficiency in Language Modeling}
The challenge of limited high-quality pretraining data (the “token crisis”) has spurred increasing interest in improving data efficiency in language models \cite{runout,wang2023datamanagement}. A line of work studies how to reuse data across multiple epochs in autoregressive (AR) models \cite{torepeat,muennighoff2025scalingdataconstrainedlanguagemodels}. For example, Muennighoff et al.\ analyze how repeated exposures to data exhibit diminishing returns, and propose extensions to the Chinchilla scaling laws that account for data reuse \cite{muennighoff2025scalingdataconstrainedlanguagemodels,hoffmann2022chinchilla}. In general, AR models show limited benefit beyond a few epochs of data repetition before overfitting or saturation occurs \cite{torepeat,muennighoff2025scalingdataconstrainedlanguagemodels}. In contrast, generative modeling communities (e.g., in vision) have long leveraged multi-epoch training and augmentation to improve generalization \cite{karras2020limited, rombach2022ldm}. However, analogous practices in language modeling remain relatively underexplored.

\subsection{Diffusion vs. Autoregressive Under Data Constraints}
A growing body of work revisits the modeling objective when data, rather than computation, is the bottleneck. Prabhudesai et al.\ systematically compare masked diffusion language models with autoregressive (AR) models under repeated passes over limited data, showing that diffusion models make better use of data repetition and surpass AR models beyond a critical compute threshold \cite{mihir}. \cite{jinjie} argues in a technical blog that diffusion language models are “super data learners” under fixed unique-token budgets, reporting a crossover where diffusion outperforms AR with extensive data reuse and discussing pitfalls in fair comparisons between the two paradigms \cite{jinjie}. Together, these sources suggest that diffusion objectives can trade extra FLOPs for improved data efficiency in data-constrained regimes.

\section{Conclusion}
Our study shows that the data efficiency of diffusion language models primarily stems from token dropout, rather than the diffusion objective itself. We further demonstrate that autoregressive models can achieve comparable or superior efficiency when equipped with stochastic regularization such as dropout and weight decay. These findings provide a unified view of multi-epoch training, suggesting that principled corruption and regularization strategies are key to mitigating the token crisis and guiding the design of future large language models.


\bibliography{references}{}
\bibliographystyle{plain}

\renewcommand{\thesubsection}{\Alph{subsection}}
\end{document}